\documentclass[runningheads]{llncs}

\usepackage{booktabs}
\usepackage{framed}
\usepackage{dirtytalk}
\usepackage{soul}
\usepackage{caption}

\usepackage[utf8]{inputenc} 
\usepackage[T1]{fontenc}    
\usepackage{hyperref}       
\usepackage{url}            
\usepackage{booktabs}       
\usepackage{amsfonts}       
\usepackage{amsmath}       
\usepackage{nicefrac}       
\usepackage{microtype}      
\usepackage{xcolor}         
\usepackage[pdftex]{graphicx}

\title{
ExaRanker:\\
Explanation-Augmented Neural Ranker
}

\author{%
   Fernando Ferraretto\inst{1} \and
   Thiago Laitz\inst{1,2} \and
   Roberto Lotufo\inst{1,2} \and\\
   Rodrigo Nogueira\inst{1,2}
}

\institute{FEEC, UNICAMP, Brazil \and
NeuralMind, Brazil}

\begin{document}

\maketitle

\begin{abstract}
Recent work has shown that inducing a large language model (LLM) to generate explanations prior to outputting an answer is an effective strategy to improve performance on a wide range of reasoning tasks. In this work, we show that neural rankers also benefit from explanations. We use LLMs such as GPT-3.5 to augment retrieval datasets with explanations and train a sequence-to-sequence ranking model to output a relevance label and an explanation for a given query-document pair. 
Our model, dubbed ExaRanker, finetuned on a few thousand examples with synthetic explanations performs on par with models finetuned on 3x more examples without explanations.
Furthermore, the ExaRanker model incurs no additional computational cost during ranking, and allows explanations to be requested on demand. Code and data are available at \url{https://github.com/unicamp-dl/ExaRanker}
\end{abstract}

\section{Introduction} \label{sec:introduction}

Pretrained transformers such as BERT~\cite{devlin-etal-2019-bert} and T5~\cite{raffel2020exploring} finetuned on hundreds of thousands of examples brought significant gains to information retrieval (IR) tasks~\cite{lin2021pretrained,nogueira2019passage,macavaney2019cedr,li2020parade,gao2021rethink,wang2021gpl,xin2021zero,hofstatter2021efficiently,formal2021splade,lu-etal-2021-less,santhanam-etal-2022-colbertv2,hofstatter2022introducing,thakur2022domain,zhuang2022rankt5}.
When queries and documents from a task of interest are similar to the ones in the finetuning data, it is likely that a model will have an effectiveness better than that of unsupervised models. For example, a monoT5~\cite{nogueira2020document} reranker finetuned on 400k positive query-passage pairs from MS MARCO~\cite{MSMARCOv3} outperforms BM25~\cite{robertson1995okapi} in 15 of 18 datasets of the BEIR benchmark~\cite{rosa2022no,rosa2022defense}.

However, when the number of labeled examples is limited, the effectiveness of the model decreases significantly. For example, a BERT reranker finetuned on a mere 10k query-relevant passage pairs performs only slightly better than BM25 on the MS MARCO passage ranking benchmark~\cite{nogueira2020document}.
Increasing the model size~\cite{rosa2022no} or pretraining it on IR-specific objectives~\cite{izacard2021contriever,gao2022cocondenser} alleviate the need for more finetuning data, at the expense of increased computational resources.

We argue that one reason neural retrievers need extensive amounts of training examples is that they are finetuned with categorical labels (e.g., true/false).
These labels provide limited information about the task to be learned, making it more challenging for the model to grasp the nuances of the task.
For example, imagine attempting to teach a human to evaluate the relevance of passages to queries, but only being able to communicate the words ``true'' or ``false'' per query-passage pair. The learning process would be more efficient if explanations in natural language were provided to explain why a passage is relevant or not to a given query.

In this work, we propose a method for training retrieval models using natural language explanations as additional labels, which reduces the need for training examples.
Furthermore, we show that few-shot LLMs such as GPT-3.5~\cite{ouyang2022training} can be effectively employed to automatically augment training examples with explanations, enabling IR practitioners to apply our method to other datasets without the need for manual annotation. Our findings indicate that the utility of incorporating explanations decreases as the number of training examples increases. Furthermore, our experimental results demonstrate that finetuning a model to generate a label prior to an explanation leads to superior performance in comparison to generating an explanation before the target label. This outcome may be counterintuitive and contradicts previous findings in chain-of-thought works\cite{magister2023teaching}.

\section{Related Work}
Recent research has shown that inducing a large language model to generate step-by-step rationales in natural language improves performance in various tasks that require reasoning~\cite{recchia2021teaching,fernandes2022learning,wang2022self,zelikman2022star,nye2022show,katz2022inferring,zhou2022least,huang2022large,bueno2022induced}.
However, experiments using induced explanations are typically carried out on models with billions of parameters, which are impractical to use in information retrieval tasks. For example, reranking 100 passages for one query using a model with 175B parameters would take at least one minute on four A100 GPUs.
In this work, we propose a method to distill the knowledge from these large models and effectively use it to improve the quality of results in a ranking task.

Our method is related to InPars~\cite{bonifacio2022inpars,inpars-v2}, Promptagator~\cite{dai2022promptagator} and UPR~\cite{sachan2022improving}, with the distinction that we augment existing target labels from training datasets with additional information relevant to the task at hand, rather than generating queries from documents. We view InPars and Promptagator as complementary to our method, and anticipate potential integration of these approaches in future research.

A large body of work addresses the task of generating explanations for a ranked list of results~\cite{singh2018interpreting,singh2018posthoc,verma2019lirme,thomas2019investigating,roy2019rex,singh2019exs,fernando2019study,sen2020curious,singh2020valid,rahimi2021explaining,zhuang2021interpretable,volske2021towards,yu2022towards,mysore2022multivector}.
For example, GenEx~\cite{rahimi2021explaining} generates noun phrases (e.g., ``impacts on medicare tax'') for a given query-document pair.
Snippet generation can also be regarded as providing explanations for why certain results are being presented to the user~\cite{tombros1998advantages,turpin2007fast,bast2014efficient,chen2020abstractive}.
A key distinction of these works in comparison to ours is that our goal is to use explanations as a means to improve the effectiveness of a retrieval model, rather than solely for the purpose of explaining a list of results to users.
We do not claim that our method makes a retriever interpretable as we have not evaluated the correctness of the generated explanations. Our goal here is orthogonal to that of building interpretable retrievers: we show that explanations improve the effectiveness of retrievers. Nevertheless, our method may help users to understand a ranked list of results, but further examination of this aspect is left for future research.

\section{Methodology}
\label{sec:method}

\begin{figure}
    \centering
    \includegraphics[scale=0.45]{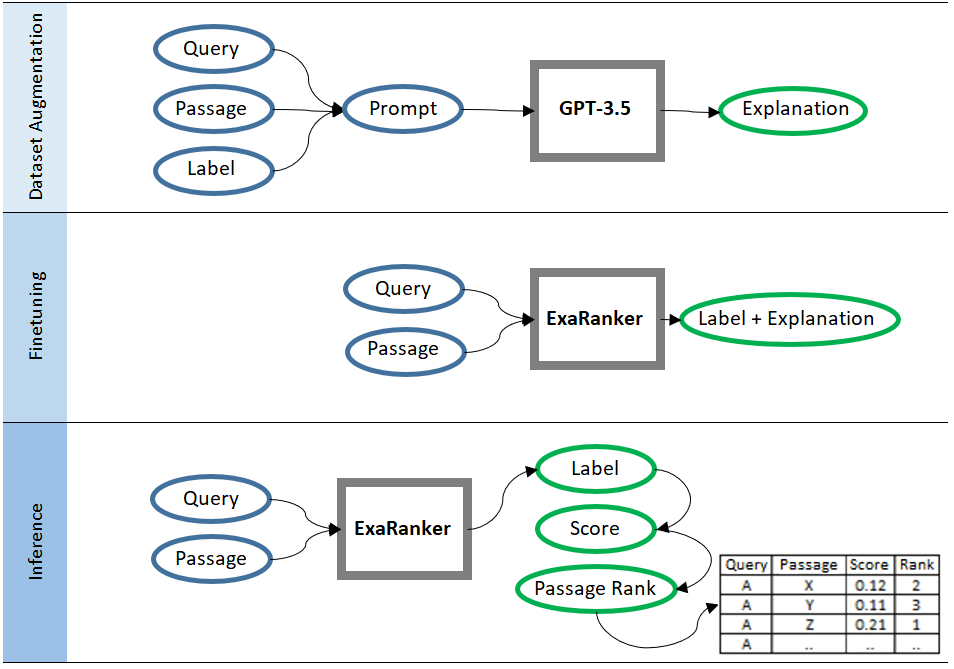}
    \caption{Method overview.}
    \label{fig:overview}
\end{figure}

The proposed method is illustrated in Figure~\ref{fig:overview}. It starts by generating explanations for query-passage-label triples using an LLM model with in-context examples. These training triples are then augmented with the generated explanations, and a sequence-to-sequence model is finetuned to produce the target label followed by the explanation. In the inference phase, the finetuned model is used to estimate the relevance of a query-passage pair, based solely on the probability assigned to the label token.

\begin{figure}
    \centering
    \includegraphics[scale=0.4]{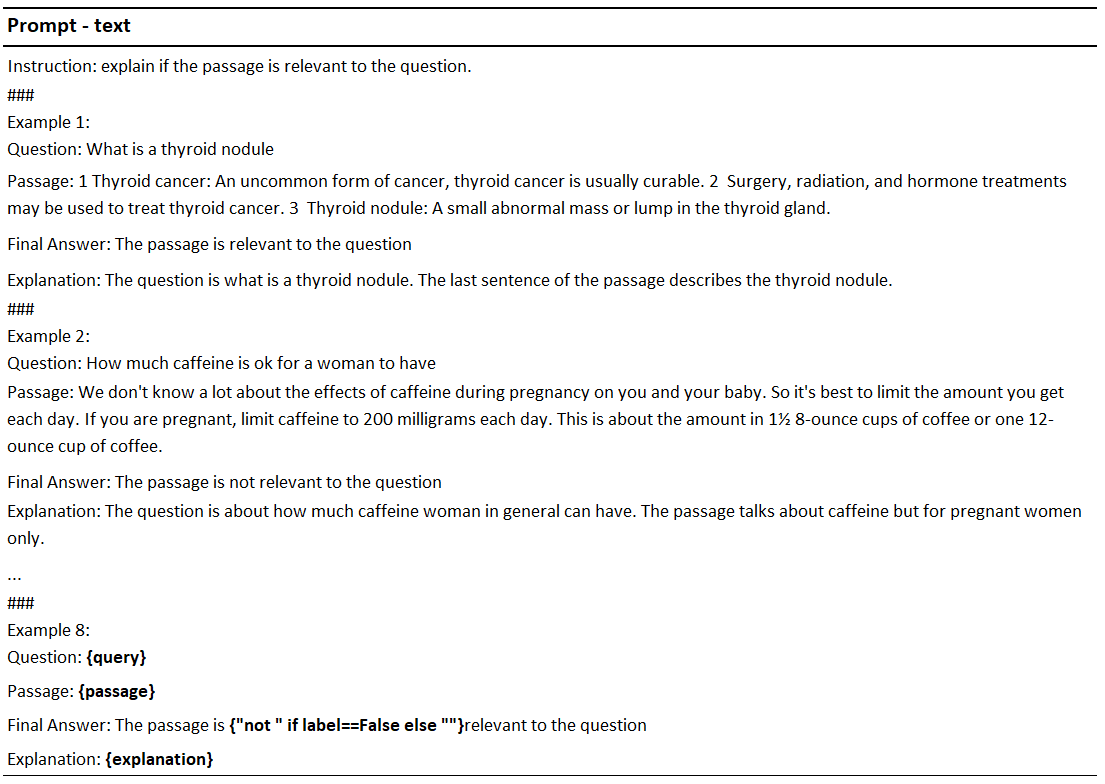}
    \caption{Prompt used to generate explanations for a query-passage-label triple (presented in Python's f-string notation).} 
    \label{fig:prompt}
\end{figure}

We begin by randomly selecting 15k pairs of (query, relevant passage) and another 15k pairs of (query, non-relevant passage) from the training set of MS MARCO passage ranking dataset, and then generate explanations for these 30k pairs. Manually generating explanations for such a large volume of text would be
cost-prohibitive, so we use \texttt{text-davinci-002}\footnote{beta.openai.com} with a few-shot prompt to infer explanations for query-passage-label triples. We use greedy decoding, and the output is limited to 256 tokens.
The few-shot prompt, illustrated in Figure~\ref{fig:prompt}, contains 7 examples that were selected from the MS MARCO training dataset, and has an average of 1400 tokens, including the 256 tokens of the explanation to be generated. As of January 2023, generating an explanation for each query-passage-label triple costs 0.028 USD using the OpenAI API. Due to cost constraints, the dataset was limited to 30k examples, which amounts to 840 USD.

Note that we also use the label (relevant or not relevant) in the prompt and request only the explanation for the triple query-passage-label. In preliminary experiments, we noticed that the model generates better explanations when the label is provided, as it is not confounded by the classification task and can focus specifically on the explanation.

Once the 30k explanations are generated, a sequence-to-sequence model is finetuned with the following input/output templates (shown using Python's f-string notation):

\noindent Input:\\
{\small \texttt{Is the question \{query\} answered by the \{passage\}? Give an explanation.}}

\noindent Output:\\
\texttt{\{label\}. Explanation: \{explanation\}}

\begin{figure}
    \centering
    \includegraphics[scale=0.35]{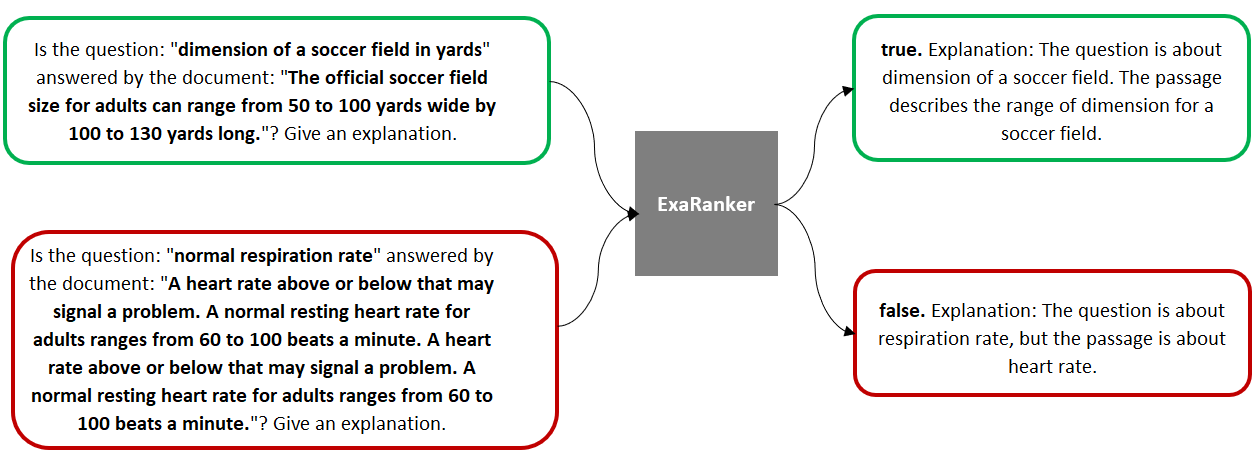}
    \caption{Illustration of input and generated outputs of a relevant (green) and non-relevant (red) query-passage pair.}
    \label{fig:template_seq2seq}
\end{figure}

The terms \texttt{\{query\}} and \texttt{\{passage\}} are the query-passage pair extracted from the MS MARCO dataset. The \texttt{\{label\}} is \texttt{true} if the passage is relevant to the query and \texttt{false} otherwise. Finally, \texttt{\{explanation\}} is the one generated by the LLM as explained above. Figure~\ref{fig:template_seq2seq} illustrates examples of input and output generated by the model.

The T5-base model was used as the starting point for the finetuning phase. The model was finetuned for 30 epochs using the AdamW optimizer~\cite{loshchilov2018decoupled} with a learning rate of 3e-5, weight decay of 0.01, and batch size of 128 examples (64 positives and 64 negatives). The maximum number of input and output tokens were each limited to 512. Sequences exceeding these values were truncated during both training and inference. For comparison purposes, a T5-base model was also finetuned using the same hyperparameters and dataset, but without the inclusion of explanations in the target text.

After finetuning, we evaluate the models on a passage ranking task. The input to the model is the same as presented before, and, as a consequence, it generates the same output pattern.\footnote{In our experiments the model always generates an output that matches the target template.} The probability of the first output token is used as the relevance score $s$ for the query-passage pair:



\[
    s= 
\begin{cases}
    1 + p_0,& \text{if } t_0 = \text{true}\\
    1 - p_0,& \text{if } t_0 = \text{false}\\         0, & \text{otherwise}
\end{cases}
\]

\noindent where $t_0$ is the token generated in the first decoding step and $p_0$ is the probability assigned by the model to that token, i.e., the probability from the softmax after the logits.

We limit the model to generate only the label for the query-passage pair (i.e., the first token) and omit the full explanation text to save processing time. Since the model was trained with causal mask, that is, only tokens generated so far influence the prediction of the next token, the relevance scores are the same had the model generated an explanation.
Alternatively, explanations can generated by decoding more tokens until the termination token (e.g., \texttt{<EOS>}) is generated.

To evaluate the real benefits of explanation in a more realistic scenario in which training data is not available, we evaluate the model in a zero-shot manner on 6 datasets from the BEIR benchmark~\cite{thakur2021beir}: Robust04~\cite{robust04}, TREC-COVID~\cite{trec-covid}, DBPedia~\cite{hasibi2017dbpedia}, FiQA~\cite{fiqa}, TREC-NEWS~\cite{soboroff2018trec} and NFCorpus~\cite{boteva2016full}. TREC-DL 2020~\cite{dl20} is used as a validation set to select the best checkpoint and hyperparameters.

\section{Results}
Main results are presented in Table~\ref{tab:main}. The first two rows are the BM25 baseline and the monoT5-base model finetuned on 400k positive query-passage pairs from the MS MARCO dataset for one epoch\footnote{\url{https://huggingface.co/castorini/monot5-base-msmarco-10k}}. We focus our analysis on the number of positive query-passage pairs as these need to be manually annotated. Thus they are an expensive component when developing a search engine. Negative query-passage pairs can be automatically selected using a retriever once queries are gathered. We also need to generate explanations for negative query-passage pairs, but we assume this cost will decrease over time with the increased availability of open-sourced LLMs.

In the third and fourth rows, we compare models finetuned using the same hyperparameters and examples, but without explanations (monoT5) and with explanations (ExaRanker). The ExaRanker model outperforms the model without explanations in all 7 datasets. On average, it improves the score by 0.8 nDCG@10 points in the zero-shot evaluation.

Compared to the BM25 baseline, the finetuned models have a much higher nDCG@10 scores. Also, ExaRanker finetuned on 15k and 10k positive examples are on average 1.7 nDCG@10 points behind monoT5-400k, which has been finetuned on 400k positive examples. This result shows the benefits of using the explanation to provide more data and reasoning during the training phase, and enable the model to succeed with much less training data.

\begin{table}[]
\centering
\begin{tabular}{lc|rrrrrrrr}
\toprule
\textbf{Model} & \textbf{Ft Pos.} & \textbf{DL 20} & \textbf{Robust} & \textbf{Covid} & \textbf{Dbp} & \textbf{FiQA} & \textbf{News} & \textbf{NFC} &
\textbf{Avg ZS}
\\
\midrule
BM25 & - & 0.478 & 0.407 & 0.594 & 0.318 & 0.236 & 0.395 & 0.321 & 0.379\\
monoT5 & 400k & 0.652 & 0.536 & 0.777 & 0.419 & 0.413 & 0.447 & 0.357 & 0.491\\
\midrule
monoT5 & 15k & 0.656 & 0.523 & 0.746 & 0.392 & 0.382 & 0.409 & 0.344 & 0.466\\
ExaRanker & 15k & 0.680 & 0.531 & 0.747 & 0.394 & 0.403 & 0.417 & 0.351 & 0.474\\
\midrule
monoT5 & 10k & 0.643 & 0.510 & 0.749 & 0.379 & 0.374 & 0.426 & 0.341 & 0.463\\
ExaRanker & 10k & 0.667 & 0.527 & 0.752 & 0.409 & 0.393 & 0.418 & 0.347 & 0.474\\
\midrule
monoT5 & 5k & 0.625 & 0.488 & 0.693 & 0.364 & 0.337 & 0.417 & 0.328 & 0.438\\
ExaRanker & 5k & 0.665 & 0.505 & 0.750 & 0.389 & 0.380 & 0.414 & 0.345 & 0.464\\
\midrule
monoT5 & 2.5k & 0.611 & 0.486 & 0.666 & 0.334 & 0.328 & 0.370 & 0.325 & 0.418\\
ExaRanker & 2.5k & 0.650 & 0.496 & 0.686 & 0.393 & 0.306 & 0.398 & 0.335 & 0.436\\
\bottomrule
\end{tabular}
\vspace{0.1cm}
\caption{Main results (nDCG@10), including average zero-shot (all except DL 20). The column ``Ft Pos.'' is the number of positive training examples on which the model was finetuned.}
\label{tab:main}
\end{table}

To better understand the benefits of using explanations, we have reproduced the training methodology with smaller datasets: 20k, 10k and 5k examples. All datasets have an equal number of positive and negative query-passage pairs. 

The ExaRanker model outperforms the model finetuned without explanations in all cases. It reinforces the evidence of the benefits of using explanations and indicates that the model benefits more when datasets are smaller. When finetuning on 15k positive examples, ExaRanker performs 0.8 points better than monoT5. When finetuned on 10k examples, the improvement is 1.1 points and increases to 2.6 when finetuned on 5k positive examples.

\begin{figure}
    \centering
    \includegraphics[scale=0.4]{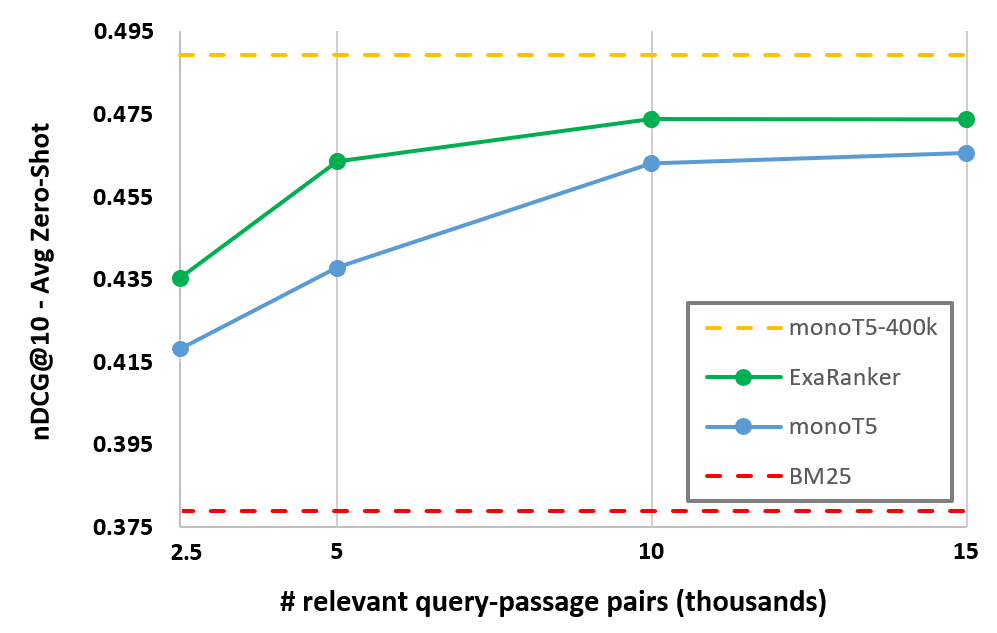}
    \caption{Average zero-shot results on 6 datasets of the BEIR benchmark when varying the number of training examples. monoT5-400k is finetuned on the 400k relevant query-passage pairs from MS MARCO without explanations.}
    \label{fig:results}
\end{figure}

Figure~\ref{fig:results} provides a visualization of this trend. We see that the improvement in using explanations decreases as the number of samples increases. However, we see a performance increase when explanations are used to augment labels compared to finetuning without explanations. Comparing the monoT5 results finetuned using 15k positive pairs with ExaRanker using 5k positive pairs, we see that the average scores are very close but ExaRanker uses 3x less data for finetuning. These results suggest that data augmentation through explanations is an effective way to transfer knowledge from larger models.

\subsection{Qualitative Analysis}
Table~\ref{tab:samples} presents some sample outputs for a qualitative comparison of the correct and incorrect predictions made by ExaRanker on the TREC-DL 2020 dataset. The model is able to generate reasonable explanations: in the second example, it was able to understand the context, correctly predicting the relevance of the passage, even though it only mentions specific terms from the query, such as ``early pregnancy''. However, it may struggle with more fine-grained relationships, such as correctly identifying the relationship between electromagnetic waves and radio waves, as shown in the third example. Overall, explanations appear to support a better understanding of the relationship between the query and the passage, both quantitatively and qualitatively.

In summary, the purpose of this method is to use the explanations to provide more information on how to perform the ranking task.
However, the models do not necessarily rely on reasoning to arrive at an answer. The explanations are used to expedite the learning process, but the most relevant score for ranking the relationship between sequences is concentrated in the label score, specifically the true or false label indicating the relevance relation, rather than in the entire explanation generated.

\begin{table}
\centering
\begin{tabular}{|p{2.5in}|p{1.5in}|p{0.5in}|}\hline
\toprule
\textbf{Input} & \textbf{Output} & \textbf{Target} \\
\midrule
Is the question: "average salary for dental hygienist in nebraska" answered by the document: "Dental Hygienist Salaries by Area. According to the Bureau of Labor Statistics, Dental Hygienists in the state of Nebraska can expect to earn an average of \$65,236 annually or \$31.36 per hour."? Give an explanation. & 
true. Explanation: The question is about average salary for dental hygienist in nebraska. The passage describes the average salary for dental hygienist in Nebraska. &
true
\\
\hline
Is the question: "can fever cause miscarriage early pregnancy?" answered by the document: "Pregnancy Early symptoms of pregnancy include a missed period, morning sickness, breast swelling, fatigue, and more. Chickenpox (varicella) Chickenpox is an illness that causes a red, itchy rash, fever, headache, sore throat, and body aches"? Give an explanation. & 
false. Explanation: The question is about if fever can cause miscarriage early pregnancy. The passage is about early symptoms of pregnancy and chickenpox. &
false
\\
\hline
Is the question: "what medium do radio waves travel through" answered by the document: "An electromagnetic wave (this includes light) does not require a physical medium - it can travel through empty space."? Give an explanation. &
false. Explanation: The question is about what medium do radio waves travel through. The passage is about electromagnetic waves and how they travel through empty space. &
true
\\
\hline
Is the question: "do google docs auto save" answered by the document: "Inter-sheet formulas in Google Docs are similar to those used in Microsoft Excel. Using this ability, you can post a value automatically in a cell on one tab that is calculate or posted in a cell on another tab."? Give an explanation. & 
true. Explanation: The question is about if google docs auto save. The passage describes how google docs auto save. &
false
\\
\bottomrule
\end{tabular}
\vspace{0.1cm}
\caption{Outputs generated by the ExaRanker model.}
\label{tab:samples}
\end{table}

\subsection{Ablation Experiments}

Two additional experiments were conducted to better understand the contribution of adding explanations as target sequences in a ranking task.

First we finetuned another ExaRanker model for 30 epochs, with the same hyperparameters but inverting the order in which label and explanations are generated: first an explanation is generated, then a relevance label. It is reasonable to expect that this order would lead to better results, as it induces the model to elaborate on the relationship between query and passage prior to predicting a label. However, as shown in Table~\ref{tab:ablation}, the nDCG@10 score dropped 9.6 points compared to generating a relevance label followed by an explanation. We believe that this behavior is due to the difficulty of using a ranking score from multiple tokens. We tried different approaches, such as using the average token probabilities of the entire sequence or using only the probability of the final token (\texttt{true} or \texttt{false}). In all cases, nDCG@10 was lower than the method described in Section~\ref{sec:method}.

\begin{table}
\centering
\begin{tabular}{lr}
\toprule
\textbf{Output template} & \textbf{nDCG@10} \\
\midrule
\texttt{\{label\}. Explanation: \{explanation\}.} & 0.680 \\
\texttt{Explanation: \{explanation\}. \{label\}. } & 0.584 \\
\bottomrule
\end{tabular}
\vspace{0.1cm}
\caption{Ablation of the output template. Results are on TREC DL 2020 and the models were finetuned on 15k positive + 15k negative examples.}
\label{tab:ablation}
\end{table}

In a second ablation study, we tested if explanations would improve a model already finetuned on a large ranking dataset. For that, we further finetuned on 15k positive + 15k negative examples with explanations a monoT5-base model already finetuned on 400k positive pairs from MS MARCO. Results in Table~\ref{tab:masmarco10k} show a negligible difference of 0.2 nDCG@10 points compared to the model finetuned only on labels. As explained before, we expect the benefits of explanations to be reduced when large training data is available. This experiment demonstrated, however, that finetuning on explanations does not negatively affect a ranker's effectiveness while equipping it with the capability of generating explanations for its relevance predictions.

\begin{table}[]
\centering
\begin{tabular}{lrrrrrrrr}
\toprule
\textbf{Model} & \textbf{DL 20} & \textbf{Robust} & \textbf{Covid} & \textbf{Dbp} & \textbf{FiQA} & \textbf{News} & \textbf{NFC} &
\textbf{Avg}
\\
\midrule
monoT5 (ft on 400k pos) & 0.652 & 0.536 & 0.777 & 0.419 & 0.413 & 0.447 & 0.357 & 0.514\\
\midrule
ExaRanker (from monoT5) & 0.701 & 0.528 & 0.756 & 0.398 & 0.406 & 0.442 & 0.352 & 0.512\\

\bottomrule
\end{tabular}
\vspace{0.1cm}
\caption{Results on finetuning Exaranker from a monoT5-base model finetuned on 400k positive examples from MS MARCO.}
\label{tab:masmarco10k}
\end{table}

\section{Conclusion}

We have proposed a method for training information retrieval (IR) models using natural language explanations as additional labels, which reduces the need for a large number of training examples. Our results demonstrate that the use of explanations can significantly improve the performance of IR models, particularly when fewer labeled examples are available.

Finally, we showed that large language models can be used to effectively generate these explanations, thus allowing our method to be applied to other IR domains and tasks. Importantly, our method does not significantly increase the time required to rerank passages as only the true/false token is used during inference. The source code and datasets used in this study are available for public use in the accompanying repository for future research and advancements of the ExaRanker method.

\section*{Acknowledgments}

This research was partially funded by grant 2022/01640-2 from Fundação de Amparo à Pesquisa do Estado de São Paulo (FAPESP).

\bibliographystyle{abbrv}

\bibliography{refs}

\end{document}